\documentclass[a4paper]{cas-dc}
\usepackage[numbers]{natbib}
\usepackage{booktabs}
\usepackage{caption}
\usepackage{fullpage}
\usepackage{color}
\usepackage {soul}
\usepackage{diagbox}
\usepackage{colortbl}
\pdfoutput=1
\usepackage{tabularray}
\definecolor{Black}{rgb}{0,0,0}

\def\tsc#1{\csdef{#1}{\textsc{\lowercase{#1}}\xspace}}
\tsc{WGM}
\tsc{QE}

\begin{document}
	\let\WriteBookmarks\relax
	\def\floatpagepagefraction{1}
	\def\textpagefraction{.001}
	\let\printorcid\relax 
	\captionsetup[figure]{labelfont={bf},labelformat={default},labelsep=period,name={Fig.}}
	\shorttitle{}    
	
	\shortauthors{C. Sun and J. Bai et al.}
	
	\title[mode = title]{Map Feature Perception Metric for Map Generation Quality Assessment and Loss Optimization}  
	
	\author[1]{Chenxing Sun}
	\author[2]{Jing Bai}
	\cormark[1]
	\cortext[1]{Corresponding author}  
	\address[1]{Laboratory of Geological Survey and Evaluation of Ministry of Education,  China University of Geosciences, Wuhan 430074, China.} 
	\address[2]{China University of Geosciences, Wuhan 430074, China.} 

	\begin{abstract}
		In intelligent cartographic generation tasks empowered by generative models, the authenticity of synthesized maps constitutes a critical determinant. Concurrently, the selection of appropriate evaluation metrics to quantify map authenticity emerges as a pivotal research challenge. Current methodologies predominantly adopt computer vision-based image assessment metrics to compute discrepancies between generated and reference maps. However, conventional visual similarity metrics—including L1, L2, SSIM, and FID—primarily operate at pixel-level comparisons, inadequately capturing cartographic global features and spatial correlations, consequently inducing semantic-structural artifacts in generated outputs. This study introduces a novel Map Feature Perception Metric (MFP) designed to evaluate global characteristics and spatial congruence between synthesized and target maps. Diverging from pixel-wise metrics, our approach extracts elemental-level deep features that comprehensively encode cartographic structural integrity and topological relationships. Experimental validation demonstrates MFP's superior capability in evaluating cartographic semantic features, with classification-enhanced implementations outperforming conventional loss functions across diverse generative frameworks. When employed as optimization objectives, our metric achieves performance gains ranging from 2\% to 50\% across multiple benchmarks compared to traditional L1/L2/SSIM baselines. This investigation concludes that explicit consideration of cartographic global attributes and spatial coherence substantially enhances generative model optimization, thereby significantly improving the geographical plausibility of synthesized maps.
	\end{abstract}

	\begin{keywords}
		cartographic generation \sep 
		map authenticity \sep 
		generative adversarial networks \sep
		evaluation metrics
	\end{keywords}
	
	\maketitle
	
	\section{Introduction}
	
	In intelligent cartography, map generation tasks driven by visual generative models face a critical challenge: enhancing synthesized map fidelity while ensuring geographical plausibility. The evaluation of cartographic authenticity constitutes a fundamental component permeating the entire generation pipeline. Contemporary approaches typically employ conditional Generative Adversarial Networks (cGANs)\citep{isola_image--image_2017} and Denoising Diffusion Probabilistic Models (DDPMs)\citep{ho_denoising_2020} to facilitate image translation from remote sensing data to cartographic representations, utilizing pixel-level metrics like L1 and MSE\cite{7797130} for quality assessment. While these error metrics partially reflect geometric correspondence, their pixel-wise computation paradigm fundamentally fails to encapsulate cartographic semantics and macrostructural integrity. Critical geographical elements including transportation networks, hydrological systems, and architectural structures remain indiscernible through such low-level measurements, with additional limitations in discerning stylometric consistency and chromatic accuracy. Furthermore, perceptual metrics like SSIM\cite{wang_image_2004} and PSNR, though incorporating human visual system considerations, demonstrate inherent inadequacies in cartographic contexts, frequently yielding misleading assessments that correlate poorly with actual map usability, thereby propagating representational inaccuracies in synthesized outputs.
	Recent advancements in evaluation methodologies have witnessed the emergence of neural-driven metrics (e.g., Fréchet Inception Distance (FID)\cite{heusel2017gans}), leveraging deep feature extractors to capture high-level semantic representations. A prevalent approach computes L2 distances between hidden layer activations in pre-trained classification networks like VGG\cite{simonyan2014very}, establishing perceptual loss functions between synthesized and reference images. Subsequent refinements by Zhang et al.\cite{zhang2018unreasonable} introduced the Learned Perceptual Image Patch Similarity (LPIPS) metric, enhancing spatial relevance through task-oriented feature calibration. While these neural-based losses demonstrate empirical effectiveness in standard image generation paradigms, their dependence on pre-trained generative architectures imposes inherent constraints, exhibiting limited adaptability to multifaceted generation challenges.
	Notably, Mustafa et al.\cite{mustafa2022training} developed a task-specific Multi-scale Discriminative Feature (MDF) loss for image restoration, demonstrating superior perceptual alignment through hierarchical feature matching. Nevertheless, such specialized solutions incur substantial retraining costs while failing to generalize across diverse cartographic generation scenarios. The advent of large-scale vision foundational models presents transformative potential, offering exceptional zero-shot generalization across downstream tasks including cartographic analysis. Contemporary studies extensively employ semantic segmentation architectures for road network extraction from satellite imagery\cite{mei_coanet_2021-1,zhou_split_2022-1}, with foundational vision models further enhancing feature discriminability in remote sensing interpretation\cite{sultan2023geosam,zhang2304text2seg}. Systematic investigations into Vision Transformer (ViT)\cite{dosovitskiy2020image}-based remote sensing classification frameworks\cite{bazi_Vision2021} have expanded methodological horizons. Building upon this, Xu et al.\cite{xu2021efficient} proposed an enhanced ViT variant addressing multi-scale processing, boundary refinement, and computational efficiency in remote sensing segmentation through hierarchical transformers and edge-aware mechanisms, achieving simultaneous improvements in segmentation precision and cartographic accuracy.
	In this work, we explore the task of map generation and the metrics commonly used for map generation. According to previous research, traditional map evaluation metrics lack consideration of feature attributes such as the geometry of map elements and spatial semantic relationships. This shows that the evaluation of generated maps and target maps is not limited to a traditional comparison between pixels but also needs to focus on the differences in the maps from the perspective of human perceptual vision. In this work, we propose our map feature perception metric based on the Vision Transfomer (ViT) model as a benchmark for evaluating map features. It also aims to evaluate the feature differences of map images in the semantic space. We show that in different generative modeling environments, our metric, which is used as the loss part of the model, generates maps that outperform the original results in terms of both qualitative and quantitative results. At the same time, our metric outperforms various metrics in the map generation task for the commonly used image quality evaluation metrics.
	We finally transformed the method into a loss function to evaluate the differences in the characteristics of map images in the semantic space. Experimental verification showed that the quality of map generation by the model was significantly improved after adopting the loss function we proposed, which further verified the effectiveness of the new index in evaluating map characteristics.

	\section{Related Work}
	
	In map generation, how to effectively use neural network models to generate high-quality map images is one of the core issues of current research. With the rapid development of deep learning technology, advanced methods such as generative adversarial networks (GANs) and their variants and generative models based on attention mechanisms have gradually been applied to map generation tasks. These methods greatly improve the accuracy and perceptual quality of generated images by introducing different network architectures and loss functions. However, existing models still face challenges in feature extraction and evaluating the quality of generated images, especially in complex mapping tasks, where accurately capturing spatial structural information is particularly important. In this paper, we briefly review the existing methods for map generation and translation tasks and explore their performance with different loss functions in the image translation task.
	\subsection{Methods of I2I}

	As I2I frameworks seek to establish inter-domain mapping relationships, the architectural representation of these cross-domain mappings critically influences generative fidelity \cite{Pang.et.al}. The objective extends beyond pixel-level replication of source imagery to encapsulating latent semantic representations and modeling intricate feature distributions. Formally, given data conforming to specific distributional constructs, generative models aim to synthesize artificial data instances that approximate the intrinsic probability distribution of authentic samples \cite{xu2015overview,oussidi_deep_2018-1}.
	The variational autoencoder (VAE) introduced by Kingma and Welling in 2013 pioneered probabilistic modeling for data generation [20], though limitations in synthesizing high-resolution outputs and inherent instability during adversarial training precipitated its eclipse by GAN-based architectures. The Pix2Pix\cite{isola2017image} framework, formulated through conditional generative adversarial networks (cGANs)\cite{mirza2014conditional}, established supervised cross-domain mapping by integrating input-output constraints, demonstrating efficacy in semantic label visualization, edge-to-object reconstruction, and chromatic synthesis. Beyond these foundational paradigms, diffusion probabilistic models (DPMs)\cite{ho_denoising_2020-1} have emerged as competitive alternatives, particularly excelling in translation tasks. Palette\cite{saharia2022palette} exemplifies this progression through a unified conditional diffusion framework addressing colorization, inpainting, and compression artifact removal without task-specific parameterization, demonstrating superior performance metrics against conventional GANs and regression baselines. Notably, StarGAN\cite{choi2018stargan} presents a paradigmatic solution for multi-domain translation via unified architecture design, while Blau et al.\cite{blau2018perception} pioneer unpaired translation methodologies, significantly broadening the technological applicability across diverse cartographic scenarios.
	\subsection{Application for the map generation task}
	Early digital mapping techniques relied heavily on manual drawing and editing processes, which were significantly time- and cost-intensive\cite{haunold_keystroke_1993}. Advances in geospatial remote sensing have promoted the application of aerial/satellite images in the field of mapping, but the feature recognition and vectorization processes still rely on manual labeling. The deep learning revolution has led to important breakthroughs in automated geographic feature extraction technology: a high-resolution image synthesis method based on conditional generative adversarial networks (Conditional GANs)\cite{isola_image--image_2017}, which introduces an adversarial loss function and a multi-scale generative-discriminative architecture to achieve end-to-end conversion of semantically labeled maps to high-fidelity map images. The Pix2Pix framework\cite{isola2017image} uses conditional adversarial networks to establish a mapping relationship between aerial images and standard-scale maps, while the GeoGenerative Adversarial Network (GeoGAN)\cite{ganguli2018geogan} innovatively integrates reconstruction loss and style transfer mechanisms to achieve intelligent generation of satellite images to standard layers. At the same time, breakthroughs in image recognition and semantic segmentation techniques have promoted the automated extraction of road networks. The DeepRoadMapper system proposed in \cite{wu2020deepdualmapper} achieves accurate reconstruction of road topology through residual network architecture and optimization of the soft intersection over union (IoU) loss function.
	\subsection{Map-generated indicators}
	The intelligent generation of map scenes is a spontaneous learning process. The results of this generation need to model the location, structure, distribution, interrelationships, and potential laws of multi-dimensional spatial entities. Furthermore, these results must conform to the cognitive framework of the real world as perceived by humans. To continuously optimize and improve the performance of the map scene generation model, it is necessary to scientifically and rationally evaluate and provide feedback on the generation results. In traditional methods for evaluating map quality, many studies rely mainly on similarity metrics such as geometric properties, directionality, and topological relationships to comprehensively evaluate the reliability of the generated map. However, these methods often ignore whether the generated map is consistent with human subjective perception. In contrast, research on image quality evaluation in the field of computer vision provides valuable ideas and can draw inspiration from the following three types of methods:
	\par
	\textbf{Subjective image quality assessment}: Subjective Image Quality Assessment: Amazon Mechanical Turk (AMT) employs a "true/false" forced-choice paradigm for subjective IQA in I2I tasks \cite{isola2017image,zhu2017unpaired}. Participants select real images from pairs containing both real and generated samples, with subsequent feedback analysis producing quantitative scores.
	\par
	\textbf{Objective image quality assessment}: Objective Image Quality Assessment: PSNR quantifies intensity differences between translated and reference images, serving as a widely adopted full-reference metric. FID measures distributional divergence between synthetic and real image datasets\cite{heusel2017gans}. Comparative studies of loss functions (L1, L2, SSIM\cite{wang_image_2004}, MS-SSIM\cite{wang_multiscale_2003}) demonstrate that combined MS-SSIM+L1 achieves superior performance across multiple quality metrics\cite{7797130}.
	\par
	\textbf{Feature perceptual loss}: Traditional metrics (L1, L2, SSIM) show limited applicability in complex tasks (e.g., super-resolution, 3D reconstruction). Prior studies demonstrate weak correlations between traditional super-resolution metrics (PSNR, SSIM) and human visual assessment\cite{johnson2016perceptual}. Perceptual loss leverages deep CNN features (e.g., VGG\cite{simonyan2014very}) to measure semantic, textural, and structural similarity, proving effective for complex image generation. Research reveals that perceptual loss efficacy stems from deep network architectures rather than pre-trained weights\cite{liu2021generic}. Comparative analyses across network architectures (untrained VGG, ImageNet-trained VGG, SqueezeNet) and training paradigms (supervised, self-supervised, unsupervised) confirm deep features' predictive capability for visual perception tasks\cite{zhang2018unreasonable}.

	\section{Methodology}
	
	Cartographic synthesis presents distinctive technical challenges as conventional image quality metrics (e.g., FID, SSIM) prove fundamentally inadequate for evaluating geospatial representations. These conventional metrics predominantly analyze pixel-wise similarity metrics, thereby inadequately capturing the intricate spatial composition and semantic richness inherent to cartographic representations. Critical geospatial elements—including transportation networks, hydrographic systems, and anthropogenic structures—require precise positional accuracy and geometrically coherent interrelationships to ensure functional validity. This necessitates the development of feature-aware evaluation frameworks that holistically assess both visual fidelity and geospatial functionality, aligning synthetic outputs with professional cartographic standards and end-user operational requirements.
	\begin{figure}
		\centering
		\includegraphics[width=\columnwidth]{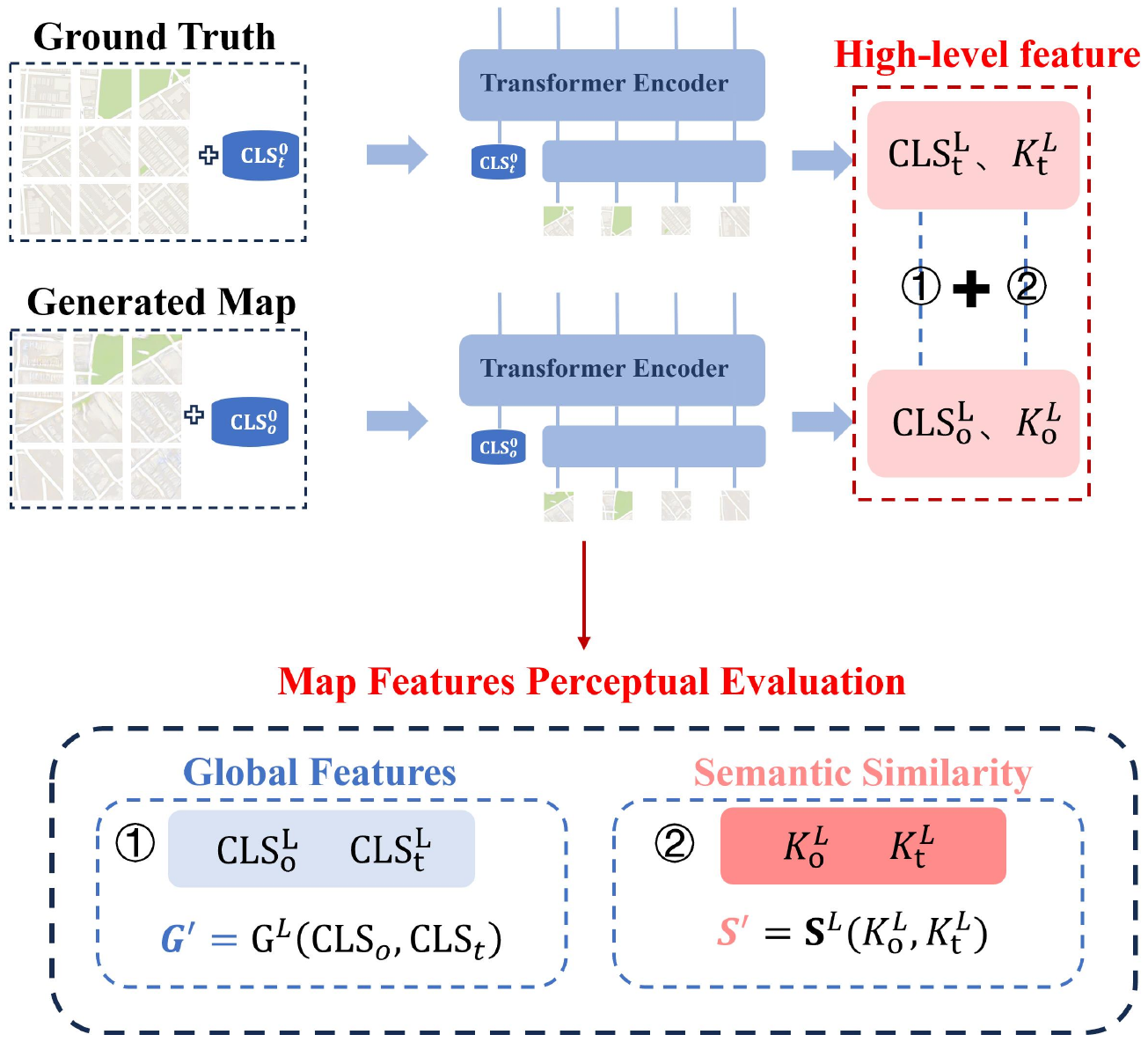}
		\caption{Overview of methods for evaluating map features. Among them, [CLS] Token is an additional learnable global feature in ViT, and $K^L$ represents the keyword of the self-attention layer QKV of the L (i.e., last) layer Block, which is used to represent the deep semantic features of the map. Finally, the global feature $G'$ and semantic similarity $S'$ of the map are calculated.}	
		\label{fig:1}
	\end{figure}
	\par
	In the image translation task, pre-trained VGG is used as an image feature extractor, and the difference in features is compared as a perceptual loss. Inspired by this idea, as shown in the Fig. \ref{fig:1}, we propose a new feature evaluation metric for maps. It combines the advantages of traditional image quality evaluation metrics (FID, SSIM, PSNR) and reasonably considers the differences between the global structural similarity and local texture characteristics of the generated map and the real map at the pixel level. At the same time, it combines the spatial correlation of map features (describing the distribution relationship of features in geographical space) and the various elements present in the map. Using the semantic performance of the ViT model as a feature extractor for the map, it is possible to obtain semantic-level contextual information in the map and better evaluate the differences between maps.
	\par
	Next, we define the map features and introduce the semantic features in the ViT model. In subsection 3.2, we describe our map feature perception metrics, and finally, in subsection 3.3, we propose the concept of map perception loss.
	\subsection{Map Feature}
	Since ViT's inception, studies have explored its capacity to encode image semantics\cite{amir2022effectiveness}:
	\par
	(1) ViT features deliver fine-grained semantic information with enhanced spatial resolution, where self-supervised features exhibit category-driven organization in classification tasks. 
	\par
	(2) Comparative analysis of ViT components (tokens, queries, values, keys) reveals superior representation in key features when correlating source-target image feature points.
	Map semantics encompass diverse elements (roads, buildings, vegetation). Current research categorizes map semantics as follows:
	\subsubsection{Building Semantic Extraction}
	ViT's self-attention mechanism enables effective building extraction in high-resolution remote sensing imagery\cite{wang_building_2022-1}. Hybrid architectures combining ViT (global context) with CNN (local details) improve earthquake-damaged building identification\cite{jia2024application}. In addition, ViT is used as a benchmark model for tasks such as extracting building outlines, classifying the semantic information of land cover types on maps, and extracting multiple semantic features (buildings, rivers, roads, vegetation, etc.) from remote sensing images\cite{zhang_extracting_2024,roy_multimodal_2023,Wang_2022}. It is worth noting that ViT's CLS token supplement semantic information is used to classify the map's features\cite{roy_multimodal_2023}.
	\subsubsection{Road Semantic Extraction}
	Roads are one of the main components of a map image. In a raster map, roads can be represented by pixel values. For example, certain pixel values can represent road areas. This method is commonly used in satellite images, remote sensing images, or rapid map generation. In remote sensing images, simple pixel-level features may not be sufficient to identify roads accurately. K. M. Kumar et al. integrated a transformer-like self-attention mechanism into the road extraction pipeline to improve the model's ability to detect long-distance dependencies and related background data\cite{kumar2024roadtransnet}. The proposed RoadTransNet model can accurately extract complex road networks from remote sensing image data by extracting complex spatial information. \cite{Fu,wang2018high} also use methods based on self-attention mechanisms to obtain contextual features of the road. R. Liu et al.\cite{Liu} evaluate the application of Transformer in road extraction in their paper. ViT processes the relationship between different regions in an image through a global self-attention mechanism, which is crucial for road extraction because roads are often narrow and continuous, and global information is required for recognition. Overall, ViT can effectively capture semantic relationships between remote regions, improving the model's performance in complex scenes.
	\subsubsection{Other Map Semantic Extraction}
	Vision transformers encode long-range dependencies between image patches via multi-head self-attention layers, generating comprehensive feature representations. Semantic features are subsequently extracted through classifier-driven softmax layers. As demonstrated in\cite{xu2021efficient}, dedicated classifiers can transform CLS tokens into semantic labels. In remote sensing applications, ViT-based architectures achieve efficient semantic segmentation through feature map fusion strategies\cite{Wang_2022}, where its encoder design enhances both processing efficiency and segmentation accuracy.
	\subsection{Map Feature Perception Metric}
\begin{figure}
		\centering
		\includegraphics[width=\linewidth]{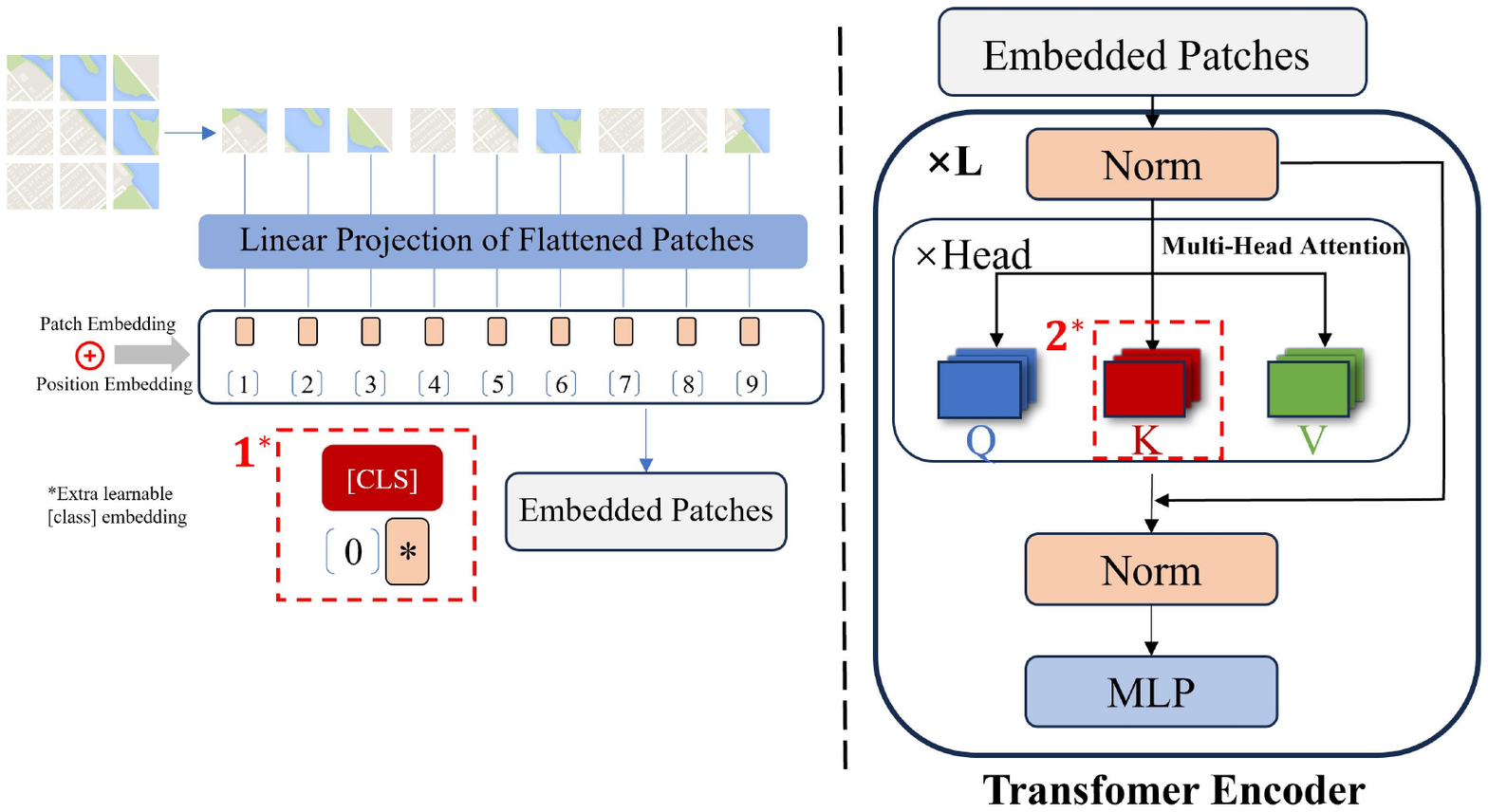}
		\caption{Forward propagation process diagram of ViT. 1*: The overall layout is used with additional learnable classification tokens.2*: ViT's multi-head self-attention mechanism after multiple propagation layers, where the tensor Key focuses on rich semantic features in each image block.}
		\label{fig:2}
	\end{figure}

	Combining the characteristics of digital maps (spatial correlation, element diversity, etc.) with the semantic information of the map, we draw on the experience from\cite{amir2021deep}, which shows that the deep features extracted in ViT can be used as dense visual descriptors. \cite{zhang_extracting_2024} Experimental verification shows that the features extracted by the self-supervised training DINO-ViT model have less noise than those of the supervised ViT model and that the deep (usually the last) layer features have local semantic information similar to that of the semantic object part. It is also pointed out in\cite{caron2021emerging} that DINO-ViT has stronger semantic capabilities and is less noisy when acquiring images. Therefore, we use DINO-ViT (self-supervised ViT) to extract deep feature information from the map. In this paper, we introduce our map feature perception metric, which evaluates map features from two aspects: global feature evaluation and map spatial correlation evaluation.See Fig. \ref{fig:2} for details.
	\par
	To better understand the rich features in the map, we follow the steps of ViT to process the image\cite{dosovitskiy2020image}, and the map I is processed into n non-overlapping image blocks. The specific treatment is as follows:
	\begin{enumerate}[(1)]
		\item Input processing:
			The input map I$\in\mathbb{R}^{H\times W\times C}$ is divided into n image blocks (patches) of size P×P,and each image block is flattened into a vector $x_p\in\mathbb{R}^{P^2\cdot C}$. Then, it is mapped to the d-dimensional space by linear embedding to obtain the embedding vector $x_p\in\mathbb{R}^d$.
		\begin{equation}
			\label{eq:1}
			x=Ex_p
		\end{equation}
		Where $E\in\mathbb{R}^{d\times(P^2\cdot C)}$ is the linear embedding matrix. The positional encoding adds a learnable [CLS] token as a global image representation.
		
		\item  Transformer Encoder:
				The input tensor z, formed by adding the position encoding, will pass through L Transformer layers. Each Transformer encoder layer consists of a multi-head self-attention (MSA) and a multi-layer perceptron (MLP) and uses layer normalization (LN) and residual connections.
				
		\begin{align}
			z_{l}^\prime=MSA\left(LN\left(z_{l-1}\right)\right)+zl-1\label{eq:2}
			\\[2pt]
			z_{l}=MLP\left(LN\left(z_{l}^\prime\right)\right)+z_{l}^\prime\label{eq:3}
		\end{align}

		where $z_l$ is the output of the layer l.		
	\end{enumerate}
	\subsubsection{Global Feature Evaluation}
	For the global features of the map, it means that the main focus is on the overall features and macrostructure of the map. For example, the map's color scheme and visual style, the type of scene it represents (city, countryside, etc.), and the semantic information on the map (such as roads, buildings, rivers, etc.). The [CLS] token, after being expressed through the model (after passing through the L-layer $z_l^0$), contains a feature classification and understanding of the map. Therefore, we reasonably use [CLS] Tokens to represent global map information. Given the generated map Io and the target map It, the corresponding $\rm CLS_o$ and $\rm CLS_t$ are obtained through the DINO-ViT model to represent the map's global information. Then, the difference between the two global features of the map is expressed by the following formula:
		\begin{align}
		Global \_{} Features=1-MSE\left({\rm CLS_o},\rm CLS_t\right) \label{eq:4}
		\\[2pt]
		MSE=\frac{1}{n}\sum_{i=1}^{n}{({CLS}_o}-{{CLS}_t)}^2\label{eq:5}
	\end{align}
	The mean square error (MSE) calculates the average error of the generated and target maps' global features. For global features, the larger the value of the global feature, the smaller the difference between the two images in terms of the characteristic elements of the scene (e.g., buildings, roads, vegetation, etc.) and the overall layout, and vice versa.
	\subsubsection{Map Spatial Similarity}
		For map spatial similarity, we need a reasonable index to evaluate the degree of similarity between the structure of a local area of the map and the overall map and to identify various spatial relationships (distance, distribution, etc.) between elements. As stated in \cite{amir2021deep}, the Transformer model demonstrates a transition in feature representation from low-level features, such as edges and textures, to high-level semantic features as the number of layers increases. In other words, ViT has captured richer semantic information in the penultimate self-attention layer, which is very important for understanding the global structure and semantic relationships of images. Therefore, we can use ViT's self-attention mechanism to capture the relationships and similarities between different regions on the map. By calculating the key (i.e., KL) of the multi-head self-attention layer, we can obtain similar information between different parts of the map:
		\begin{align}
			Spatial\_{}Similarity=\Vert S^L(I_o)-S^L(I_t) \Vert_F\label{eq:6}
			\\[2pt]
			S=cos\_{}sim\left(k_i^L\left(I\right)-k_j^L\left(I\right)\right)\label{eq:7}
			\\[2pt]
			K^L=X^LW_K^L \label{eq:8}
		\end{align}
		\par
		Among them, L is the last layer of self-attention, $\rm K^L$ is the spatial feature key of the last layer, $\rm X^L$ is the input of the L layer, and $ \rm W_K^L$  is the weight matrix of the L layer key. The cosine similarity is used to obtain the self-similarity matrix $\rm S\in\mathbb{R}^{\left(n+1\right)\times(n+1)}$ of the key, where n is the number of patches after the image is divided.
		\par
		In the map's self-similarity descriptor S, the local texture and the surrounding spatial layout, shape, and perceptual semantics can be well retained while discarding the map's appearance information. This has been proven effective in previous studies\cite{sun2025c2gm}. We chose the key's self-similarity to represent the map's spatial similarity instead of using query and value\cite{amir2021deep}.
	\subsection{Map Feature Perception Loss}
	Based on the map feature perception metric we proposed in subsection 3.3, we know that our understanding of the map is divided into two parts, so our objective function is transformed into the following, divided into two parts: the loss of global features and the loss of map spatial semantics, which can be expressed as:
	\begin{equation}
		\mathcal{L}=\lambda_1L_G{+\lambda}_2L_S\label{eq:9}
	\end{equation}
	where $L_G=G_{\left[CLS\right]}^L\left(I_o\right)-G_{\left[CLS\right]}^L\left(I_t\right)$is denoted as the global feature loss (see Eq. (\ref{eq:4})), $\rm L_S$ is expressed as the spatial semantic loss(see Eq. (\ref{eq:8})),and $\lambda_{1\ }$and $\lambda_2$ are the hyperparameters of the two. We set\ $\lambda_1=10$, $\lambda_2=1$ based on the experimental settings.
	\par
	It should be noted that the loss is unstable only on features extracted from the deep network\cite{blau2018perception}. Due to pooling in the network's hidden layers, there is no guarantee that each input has a unique potential representation. Feature loss is, therefore, often combined with a regularization term (e.g., the L2 or L1 norm), and the weights of each loss component need to be carefully tuned. In this paper, our loss also incorporates the L1 loss, which is uniformly referred to as the map feature perception loss. It will not be described in detail here. 
	\section{Experimental evaluation and analysis}
	In this section, we perform image translation on different map datasets to verify the effectiveness of our metrics in the map translation task. Below, we first introduce the details of the experiments, such as the datasets, metrics, and selected models. In subsection 4.2 and subsection 4.3, we design qualitative and quantitative comparison experiments to evaluate different methods, compare metrics, and analyze the effect of different loss functions on the quality of generated maps. Further, subsection 4.4 contains our analysis of the effectiveness of map perception metrics. Finally, subsection 4.5 and subsection 4.6 provide an in-depth interpretation of our metrics.
	
	\subsection{Experimental Details}
	\subsubsection{Datasets}
\begin{figure}[htpb]
		\centering
		\includegraphics[width=\linewidth]{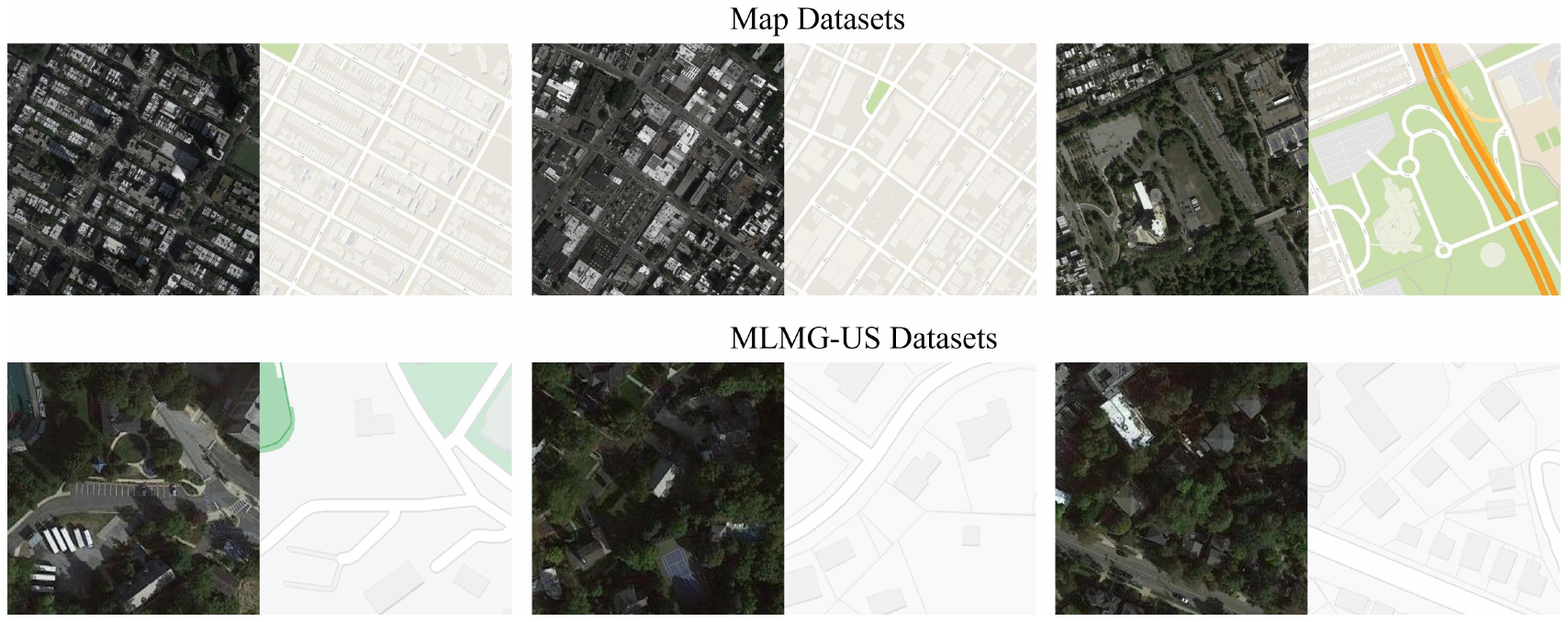}
		\caption{Map datasets, MLMG-US datasets, some paired aerial images (left) and map images (right).}
		\label{fig:3}
	\end{figure}

	To comprehensively verify the proposed map evaluation index's usability and effectiveness, multiple representative datasets were used for experimentation. As shown in Fig. \ref{fig:3}, We selected the following two datasets about previous studies\cite{isola2017image,Fu}: 
	\par
	(1) Maps datasets: consists of Google online maps, including high-resolution aerial images and corresponding map-matching samples.
	\par
	(2) MLMG-US datasets: contains paired samples of aerial images and maps of US cities.
	These datasets consist of training and test sets from different areas of the same city, ensuring consistency and representativeness of the data. Each dataset's image size is 256 × 256 pixels.
	\subsubsection{Model Selection}
	As shown in Table \ref{tab:1}, the experimental model is set up as follows: We collected several popular generative models today, divided them into the GANs category and the currently popular diffusion model category, and experimented with some metrics to obtain results. At the same time, tests were conducted on different map datasets.
	\begin{table}[]\rmfamily
		\centering
		\caption{Selection of different generative models and indicators}
		\label{tab:1}
		\begin{tabular}{cccc}
			\toprule
			Type                  & Model     & Metric(base)  & Metric(ours) \\
			\midrule
			\multirow{5}{*}{GANs} & Pix2pix   & FID SSIM PSNR & MFP          \\
			\cmidrule(r){2-4}
			& Pix2pixHD\cite{wang2018high} & FID SSIM PSNR & MFP          \\
			\cmidrule(r){2-4}
			& CycleGAN\cite{zhu2017unpaired}  & FID PSNR      & MFP          \\
			\cmidrule(r){2-4}
			& TSIT\cite{jiang2020tsit}      & FID PSNR      & MFP          \\
			\cmidrule(r){2-4}
			& SMAPGAN\cite{chen_smapgan_2021-1}   & FID PSNR      & MFP          \\
			\cmidrule(r){2-4}
			\multirow{2}{*}{DMs}  & Atme\cite{solano2023look}      & FID SSIM PSNR & MFP          \\
			\cmidrule(r){2-4}
			& C2GM      & FID SSIM PSNR & MFP          \\ 
			\bottomrule
		\end{tabular}
	\end{table}
	\subsubsection{Evaluation Metrics}
	A multifaceted approach was employed to comprehensively evaluate the quality of the generated maps, incorporating a combination of quantitative and qualitative analysis methods. Specifically, we selected the evaluation metrics SSIM, FID, and PSNR, and the new metric we proposed in subsection 3.2, specifically designed to evaluate features unique to maps. These metrics were chosen because they evaluate the quality of the generated maps from different perspectives: SSIM focuses on structural similarity, FID evaluates the overall distribution, PSNR measures differences at the pixel level, while our proposed metric for map feature perception focuses on attributes specific to maps. To ensure the reliability of the assessment, we calculate these indicators for each generated map sample and calculate the average value.
	\subsubsection{Experimental Setup}
	We designed two sets of experiments, one on the Maps dataset using different models for training and validation and the other on the MLMG-US dataset. Both sets used the same number of training rounds, i.e., 200. All models used the Adam optimizer, where it was set to 0.5 and was set to 0.999. The learning rates are different. The Maps group's learning rate is 2 × $10^{-4}$. The MLMG-US group's learning rate is 1 × $10^{-4}$. It is worth noting that the Atme model is prone to mode collapse. After testing, training the model with a learning rate of 1 × $10^{-5}$ is more appropriate. Regarding the loss function, the two groups were subdivided into two experiments: one group utilized the original loss function of each model, while the other group incorporated the map perception loss proposed to the original loss function. This design enables a direct comparison of the impact of the map perception loss on model performance. All experiments were performed on a training platform equipped with two Tesla 4 GPUs to ensure consistency of computing resources.
	\par
	In addition, we compare our map feature perception loss with the most widely used loss functions, such as L1, L2, etc., including the perceptual loss LPIPS\cite{zhang2018unreasonable}. We use the classic generative model pix2pix as the baseline and train it for 200 rounds in MLMG-US. The remaining learning rate configurations correspond to those above.
	\par
	Through this multi-faceted and multi-perspective verification method, we aim to comprehensively evaluate the usability and effectiveness of the proposed map evaluation indicators. This proves the universality of our method and reveals performance characteristics in different geographical environments and map styles, providing valuable insights for future improvements.
	\subsection{Qualitative Comparison of Experiments}
	\begin{figure*}[p]
		\centering
		\includegraphics[width=\textwidth,height=\textheight,keepaspectratio]{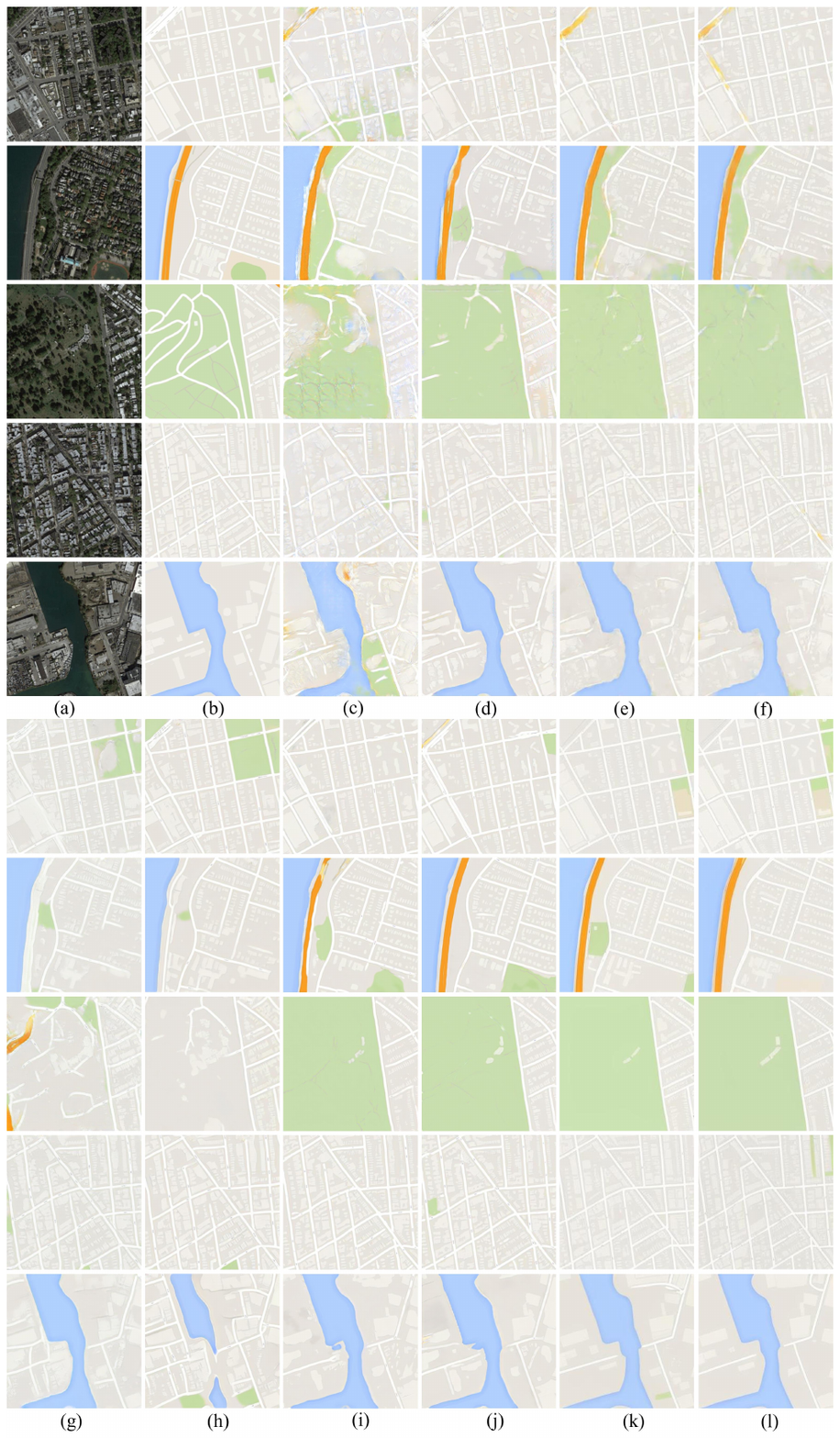}
		\caption{Qualitative results of each model on the Maps datasets using different methods. The asterisk (*) denotes model results with the additional incorporation of our map feature loss function. (a)input. (b)GT. (c)Pix2pix. (d)Pix2pix*. (e)Pix2pixHD. (f)Pix2pixHD*. (g)CycleGAN. (h)CycleGAN*. (i)ATME. (j)ATME*. (k)C2GM. (l)C2GM*.}
		\label{fig:4}
	\end{figure*}
	
	\begin{figure*}[p]
		\centering		\includegraphics[width=\textwidth,height=\textheight,keepaspectratio]{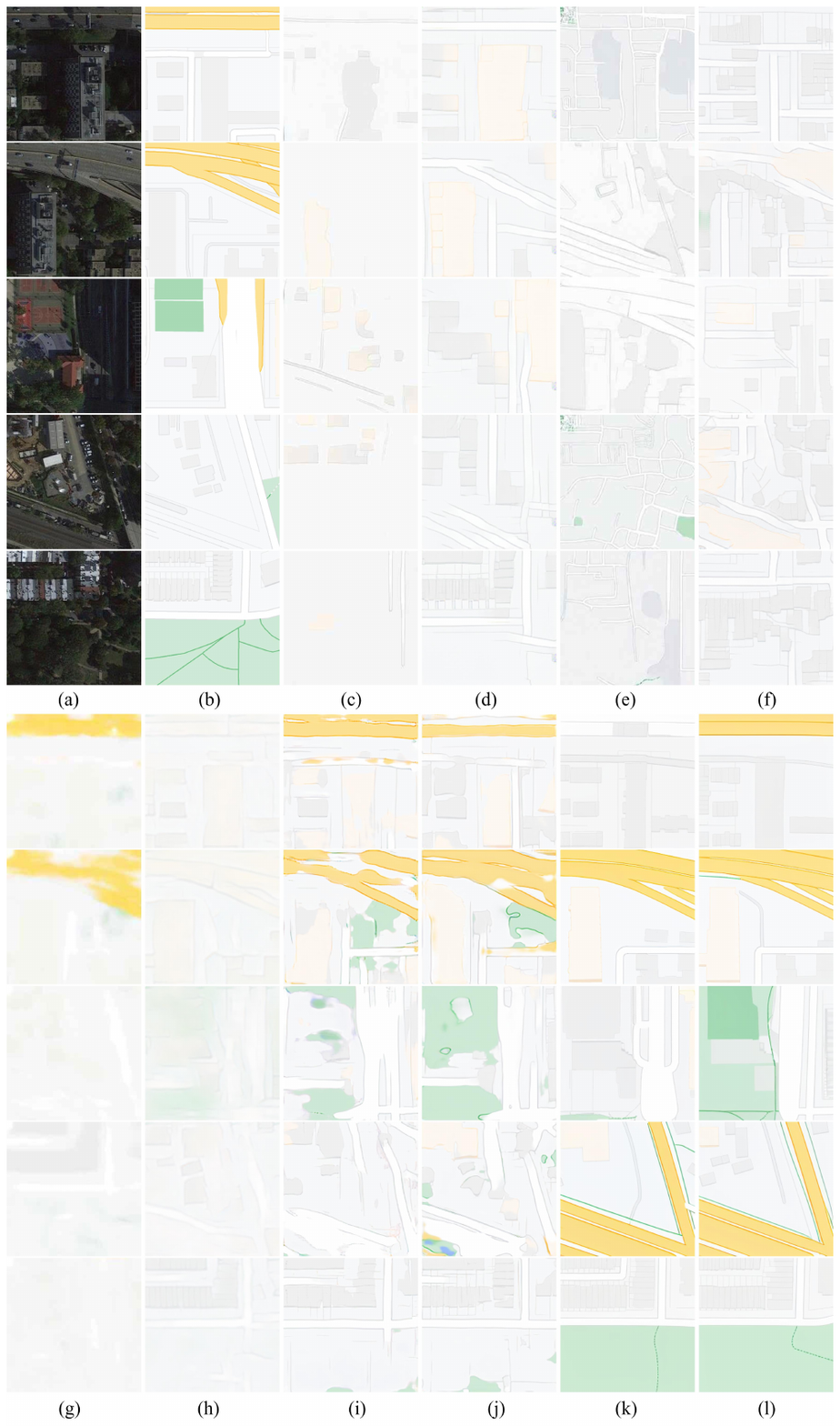}
		\caption{Qualitative results of each model on the US-MLMG datasets using different methods. The asterisk (*) denotes model results with the additional incorporation of our map feature loss function. (a)Input. (b)GT. (c)Pix2pix. (d) Pix2pix*. (e)CycleGAN. (f) CycleGAN*. (g)SMAPGAN. (h) SMAPGAN*. (i)ATME. (j)ATME*. (k)C2GM. (l)C2GM*.}
		\label{fig:5}
		
	\end{figure*}
	
\begin{figure*}[p]
		\centering
		\includegraphics[width=\textwidth,height=\textheight,keepaspectratio]{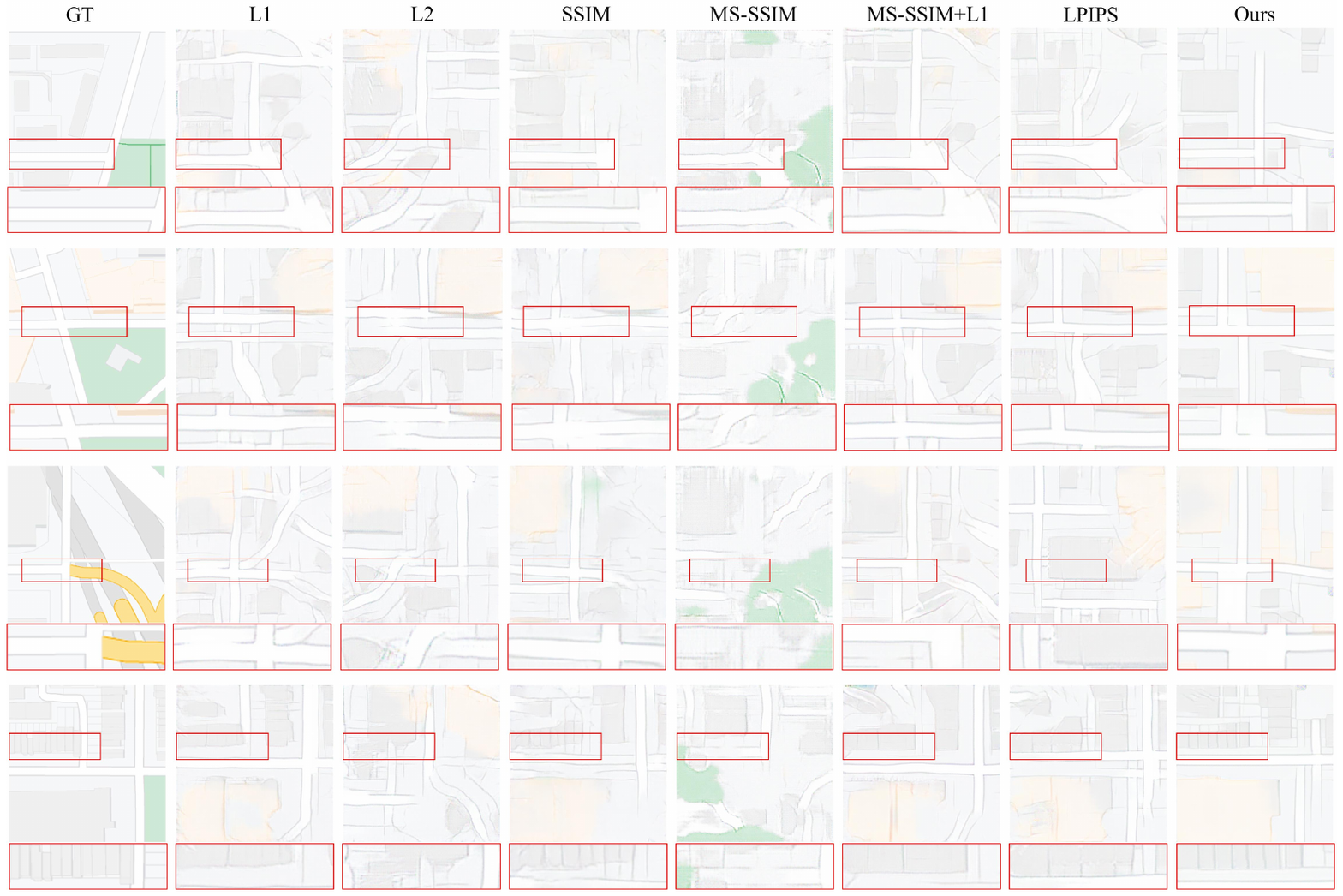}
		\caption{Results of inference using different loss functions after training the pix2pix model. Our loss improves the continuity of the roads (first to third rows), and the buildings are more evenly distributed in space (fourth row).}
		\label{fig:6}
	\end{figure*}

	First, we studied different methods on the Maps and MLMG-US datasets. Fig. \ref{fig:4} shows the results of each model on the Maps data (including GAN-based and diffusion-based methods). Our figure shows that by adding a map feature perception loss, the model can more accurately identify map elements (building groups, infrastructure, water areas, etc.) and clarify the road structure. Fig. \ref{fig:5} shows the results of each model on the US-MLMG dataset using different methods. Compared with the group that only uses the original loss function, after adding our loss function transformed by the map feature perception metric (as in the second row, the completeness of the road, bridge, and building clusters is higher), this shows that our method correctly focuses on the semantic matching regions between the output image and the real image, and also determines the spatial relationship of semantic features in the image. To further demonstrate the advantages of our approach, a comparison is made for each commonly used loss function against a basic model, pix2pix. As seen in Fig. \ref{fig:6}, our method makes the overall road structure more continuous on some road structures and, at the same time, more accurately delineates the boundaries of map semantic features, such as rivers, grasslands, and other semantic features.
	
	\subsection{Quantitative Comparison of Experiments}
	The results of the experiments are shown in Tables \ref{tab:2}, \ref{tab:3}, \ref{tab:4}, and \ref{tab:5}. The findings indicate that incorporating the MFP loss function into GAN-type models leads to enhancements in various metrics across both datasets, suggesting that the generated maps' quality, authenticity, and structural integrity have been substantially elevated. It is worth noting that the quantitative assessment results using our map quality assessment indicator MFP are consistent with the qualitative analysis in Section 4.2. High MFP values usually accompany high-quality maps and rank high in other evaluation indicators.
	The diffusion model generally outperforms GAN-based models in all metrics. Although the improvement in each indicator after adding the MFP loss is relatively limited, it may be because the model can already effectively capture the semantic features in the map, showing strong generative ability. However, although the improvement in the indicators is not significant, by comparing the results generated in Figs.\ref{fig:3} to \ref{fig:6}, it is clear that our method is superior to similar models using traditional methods in terms of map quality. Specifically, our method not only effectively improves the semantic information perception ability of the generated map but also makes the generated map more similar in characteristics to the real map. This shows that after adding the map feature loss, the model can more accurately consider the semantic structure of the map during the generation process, thereby further improving the quality and authenticity of the generated results.
	\par
\begin{figure}[htpb]
		\centering
		\includegraphics[width=\linewidth]{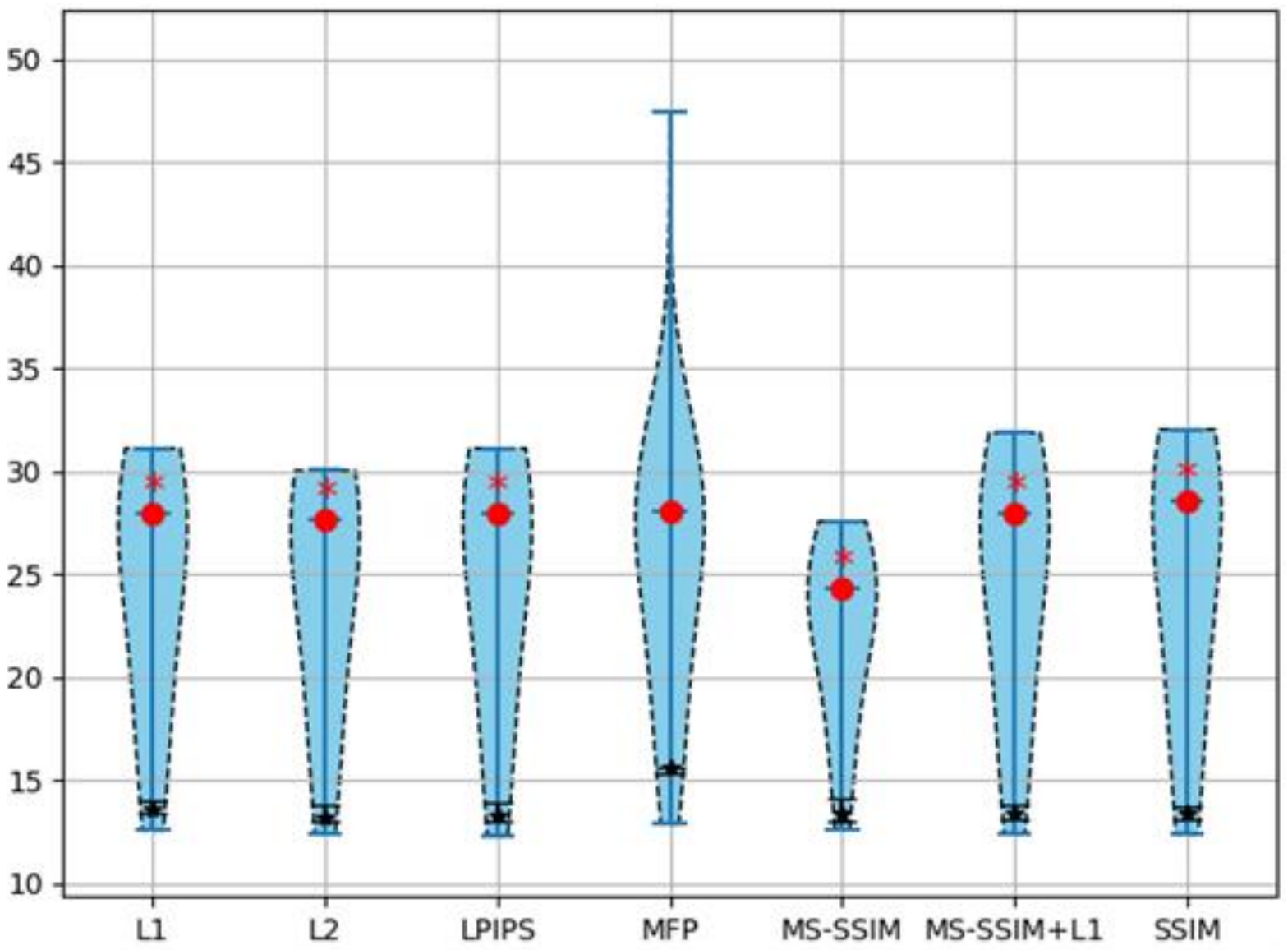}
		\caption{Violin plots were used to illustrate the distribution of PSNR↑with different loss functions on the MLMG-US dataset. The red points show the mean values. The black color indicates the 95\% confidence interval for the 5th percentile. Red asterisks indicate one-tailed t-tests for the means giving statistically significant differences at. As can be seen in the figure, our method generates more high-quality map images as well as generates less low-quality map images.}
		\label{fig:7}
	\end{figure}

	To further investigate the validity of our metrics, we compared them with different loss functions in the same experimental model. As illustrated in Table \ref{tab:6}, a series of experiments have been conducted employing a variety of widely utilized loss functions, including L1, L2, SSIM, MS-SSIM, and MS-SSIM, on the MLMG-US dataset. Our method generally outperforms other methods regarding the mean of multiple indicators. The impact of specific loss functions on metric evaluation is disregarded (e.g., employing SSIM as a loss may result in elevated SSIM metric evaluation values). Utilizing PSNR as an objective metric, our method exhibits clear advantages for both the worst and best results, as demonstrated in Fig. \ref{fig:7} (as evident from the blue dots and distribution in the figure).
\begin{table}\rmfamily
	\arrayrulecolor{black}
	\caption{Evaluation results of GANs models under different metrics for the Maps dataset using different methods}
	\label{tab:2}
	\resizebox{\linewidth}{!}{	
		\begin{tabular}{@{}llllll@{}} 
			\toprule
			Method                             & Model     & FID↓           & SSIM↑           & PSNR↑          & MFP↑            \\ 
			\midrule
			\multirow{3}{*}{Original
				Loss}   & Pix2pix   & 182.24         & 0.5675          & 23.76          & 0.4786          \\
			& Pix2pixHD & 129.90         & 0.7258          & 21.24          & 0.71            \\
			& CycleGAN  & 123.62         & 0.6641          & 24.93          & 0.3967          \\ 
			\cmidrule{1-6}
			\multirow{3}{*}{\textbf{+MFPloss}} & Pix2pix   & 123.32         & 0.6915          & \textbf{25.96} & 0.6798          \\
			& Pix2pixHD & 126.12         & \textbf{0.7361} & 21.43          & 0.7321          \\
			& CycleGAN  & \textbf{72.44} & 0.7050          & 24.98          & \textbf{0.778}  \\
			\bottomrule
	\end{tabular}}

\end{table}

\begin{table}\rmfamily
	\centering
	\caption{Evaluation results of different models under different metrics for the Diffusions class of the Maps dataset using different methods}
	\arrayrulecolor{black}
	\label{tab:3}
	\resizebox{\linewidth}{!}{
		\begin{tabular}{llllll} 
			\toprule
			Method                             & Model & FID↓           & SSIM↑           & PSNR↑          & MFP↑             \\ 
			\midrule
			\multirow{2}{*}{Original
				Loss}   & Atme  & 49.86          & 0.7591          & 27.10          & 0.8001           \\
			& C2GM  & 48.17          & 0.70            & 26.55          & 0.8001           \\ 
			\cmidrule{1-6}
			\multirow{2}{*}{\textbf{+MFPloss}} & Atme  & 46.56          & \textbf{0.7619} & \textbf{27.27} & \textbf{0.8256}  \\
			& C2GM  & \textbf{44.51} & 0.7084          & 26.55          & 0.8018           \\
			\bottomrule
	\end{tabular}}
	\arrayrulecolor{black}
\end{table}

\begin{table}\rmfamily
	\centering
	\arrayrulecolor{black}
	\caption{Evaluation results of each model in the GANs category of the MLMG-US dataset under different metrics using different methods}
	\label{tab:4}
	\resizebox{\linewidth}{!}{
		\begin{tabular}{llll} 
			\toprule
			Method                             & Model    & FID↓            & PSNR↑           \\ 
			\midrule
			\multirow{4}{*}{Original
				Loss}   & Pix2pix  & 138.45          & 24.46           \\
			& CycleGAN & 138.92          & 22.52           \\
			& TSIT     & 123.45          & 24.17           \\
			& SMAPGAN  & 246.88          & 27.01           \\ 
			\cmidrule{1-4}
			\multirow{4}{*}{\textbf{+MFPloss}} & Pix2pix  & 132.62          & 25.10           \\
			& CycleGAN & 109.57          & 23.89           \\
			& TSIT     & \textbf{109.25} & 24.87           \\
			& SMAPGAN  & 178.29          & \textbf{27.17}  \\
			\bottomrule
		\end{tabular}
	}
	\arrayrulecolor{black}
\end{table}

\begin{table}\rmfamily
	\centering
	\arrayrulecolor{black}
	\caption{Evaluation results of each model in the DMs class of the MLMG-US dataset under different metrics using different methods}
	\label{tab:5}
	\resizebox{\linewidth}{!}{
			\begin{tabular}{llllll} 
			\toprule
			Method                             & Model & FID↓           & SSIM↑           & PSNR↑          & MFP↑           \\ 
			\midrule
			\multirow{2}{*}{Original
				Loss}   & Atme  & 139.61         & 0.7306          & \textbf{25.50} & 0.465          \\
			& C2GM  & 58.38          & 0.7501          & 21.23          & 0.60           \\ 
			\cmidrule{1-6}
			\multirow{2}{*}{\textbf{+MFPloss}} & Atme  & 101.99         & 0.7454          & 25.03          & \textbf{0.66}  \\
			& C2GM  & \textbf{54.62} & \textbf{0.7512} & 21.32          & 0.63           \\
			\bottomrule
		\end{tabular}
	}
	\arrayrulecolor{black}
\end{table}

\begin{table}\rmfamily
	\centering
	\arrayrulecolor{black}
	\caption{Evaluation results based on different metrics using different loss functions based on the Pix2pix model}
	\label{tab:6}
	\resizebox{\linewidth}{!}{
		\begin{tabular}{lllll} 
			\toprule
			Method     & FID↓            & SSIM↑           & PSNR↑         & MFP↑           \\ 
			\midrule
			L1         & 194.18          & 0.7001          & 24.46         & 0.275          \\
			L2         & 214.47          & 0.6808          & 24.26         & 0.286          \\
			SSIM       & 210.71          & \textbf{0.7482} & 24.53         & 0.275          \\
			MS-SSIM    & 244.52          & 0.6503          & 22.23         & 0.183          \\
			MS-SSIM+L1 & 186.53          & 0.7             & 24.53         & 0.297          \\
			LPIPS      & 179.23          & 0.7012          & 24.53         & 0.336          \\
			MFP        & \textbf{132.62} & 0.7174          & \textbf{25.1} & \textbf{0.61}  \\
			\bottomrule
		\end{tabular}
	}
	\arrayrulecolor{black}
\end{table}
	\begin{figure*}[!h]
	\centering
	\includegraphics[width=\textwidth,height=\textheight,keepaspectratio]{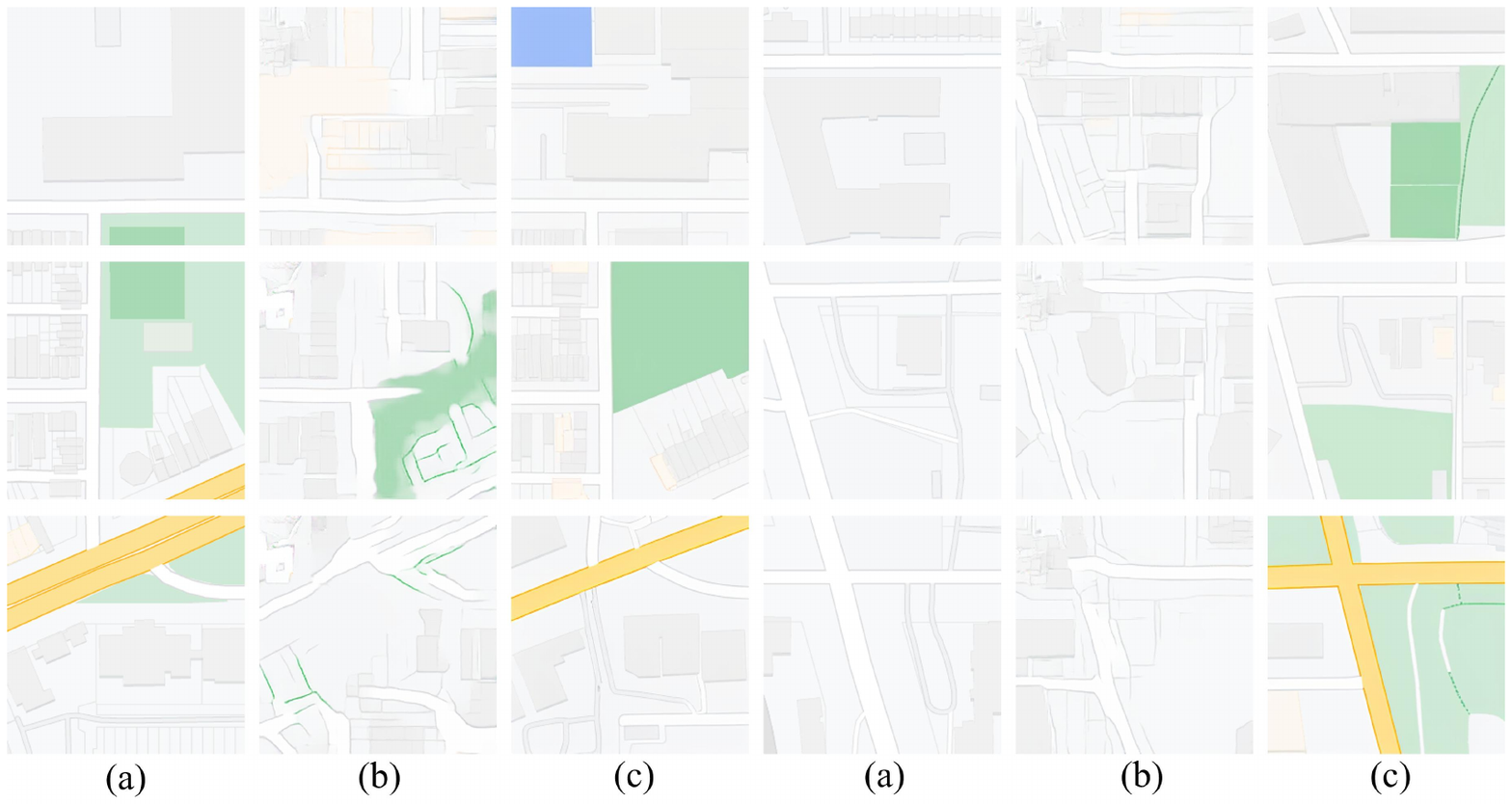}
	\caption{Sample comparison of the output of the two map generation models. (a) GT. (b) TSIT. (c) C2GM.}
	\label{fig:8}
\end{figure*}
\begin{table}\rmfamily
	\centering
	\arrayrulecolor{black}
	\caption{Comparison results of two groups of indicators of different model generation effects}
	\label{tab:7}
	\resizebox{\linewidth}{!}{
			\begin{tabular}{lllll} 
			\toprule
			\diagbox{Model}{Metric} & FID↓   & SSIM↑  & PSNR↑ & MFP↑    \\ 
			\midrule
			TSIT                    & 109.25 & 0.7142 & 24.87 & 0.5910  \\
			C2GM                    & 31.60  & 0.7503 & 24.79 & 0.7351  \\
			\bottomrule
		\end{tabular}
	}

	\arrayrulecolor{black}
\end{table}

	\subsection{Map Perception Metric Validity Analysis}

	This subsection aims to evaluate the effectiveness of the proposed map feature perception metric. To this end, the index is evaluated using traditional metrics (e.g., SSIM and FID) and our proposed index. This evaluation is performed through a qualitative and quantitative comparison of maps generated with different effects. The aim is to verify the accuracy and robustness of the index in reflecting the quality of map generation. We selected two models that performed well in the GANs and DFs categories: TSIT and C2GM. C2GM is a new framework for generating multi-scale tile maps based on the conditional diffusion model, currently the best-performing generative model. Naturally, we use TSIT as a group that produces poor results and C2GM as a group that produces good results. In principle, for the map perception index under consideration, the value of MFP for maps exhibiting poor generation effects will be lower, while the value of the natural index for maps demonstrating good generation effects will exceed that of the low group.
	\par
	Our results, shown in Table \ref{tab:7} and Fig. \ref{fig:8}, clearly indicate that the C2GM method, which generates a better map, has an MFP value of 0.7351, an FID of 31.60, an SSIM of 0.7503, and a PSNR of 24.79. The MFP value of C2GM is significantly higher than that of TSIT, and the performance of FID and SSIM is also more satisfactory. The lower FID value indicates that the map generated by C2GM has a high similarity to the feature space of the real map. SSIM and PSNR also show a high structural and pixel similarity, indicating a good generation effect. In the TSIT group, although SSIM and PSNR show some quality, the FID value is high (close to 109), indicating that the distribution of the generated map in the overall feature space is quite different from that of the real map. It can be seen that our metric (MFP) does take into account the spatial structure and semantic features of the map. At the same time, compared with traditional metrics such as SSIM and FID, MFP can better distinguish maps generated with different effects and is consistent with visual effects and the performance of other metrics, which further verifies its effectiveness in map generation evaluation.
	\begin{figure}[!h]
		\centering
		\includegraphics[width=\linewidth]{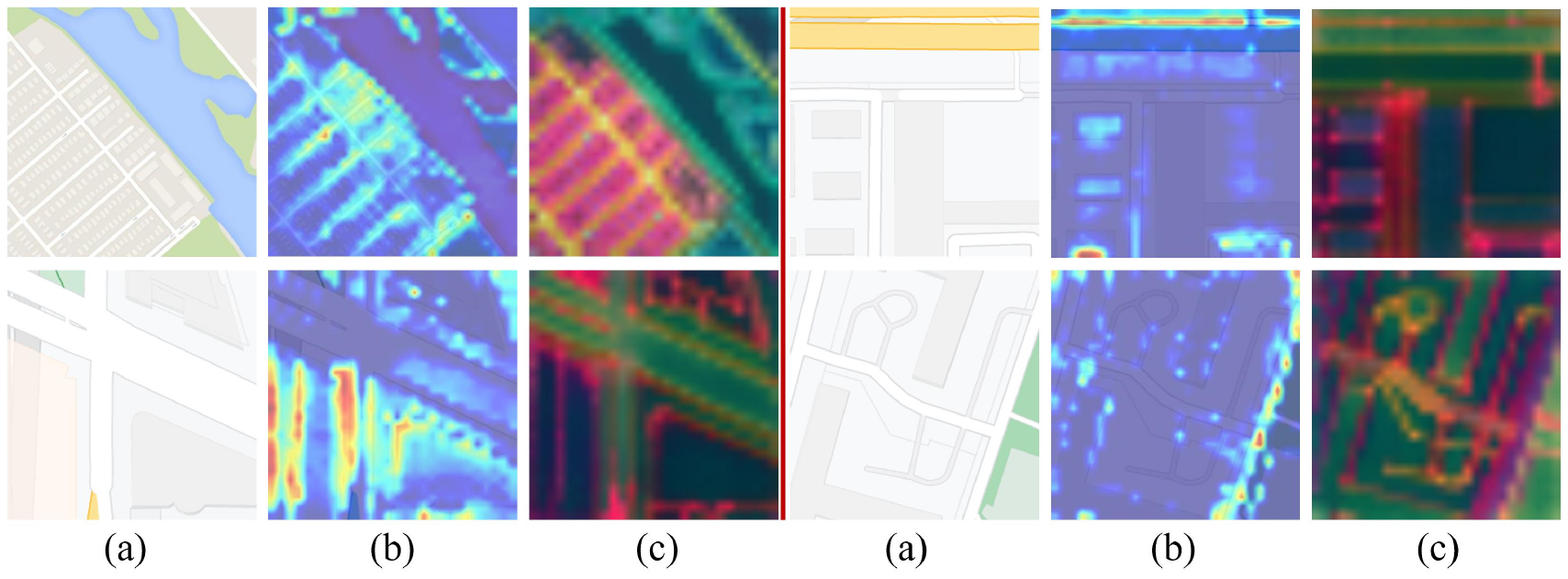}
		\caption{The first row shows an example of the Map datasets, and the second row shows an example of the MLMG-US datasets. (a) shows the original image. (b) shows the attention visualization. The darker the color, the greater the weight of the place the model focuses on. (c) shows the PCA visualization.}
		\label{fig:9}
	\end{figure}
	\subsection{Research on Semantic Visualization of Map Features}
	Next, to gain a deeper understanding of our map feature perception metric, which focuses on certain features, we visualize the index when extracting the semantic meaning of the map. Specifically, we use each map as input and the Grad-Cam method\cite{selvaraju2017grad} to obtain the output feature map of the last layer and its gradient by forward and backward propagation of ViT. We obtain the activation map by weighted superposition, activation, and normalization of the feature map, which is mapped to the input image to achieve attention localization. The visualization result is shown in Fig. \ref{fig:9} (b). We can see that the model pays more attention to the edges of some map elements. Finally, we visualize the map spatial similarity in the metric and perform a PCA dimensionality reduction on the self-similarity descriptor of $\rm K^L$. Fig. \ref{fig:9} (c) does not focus on the appearance of the map (such as the arrangement of buildings, the color of roads, etc.) but only captures the main semantic information of the map (the same semantic parts are processed with the same color). The results of the visualization processing further illustrate that our index can effectively capture the spatial semantic information in the map.
	\subsection{Correlation of Map Feature Perception Indicators Research}
\begin{figure}[!h]
		\centering
		\includegraphics[width=\linewidth]{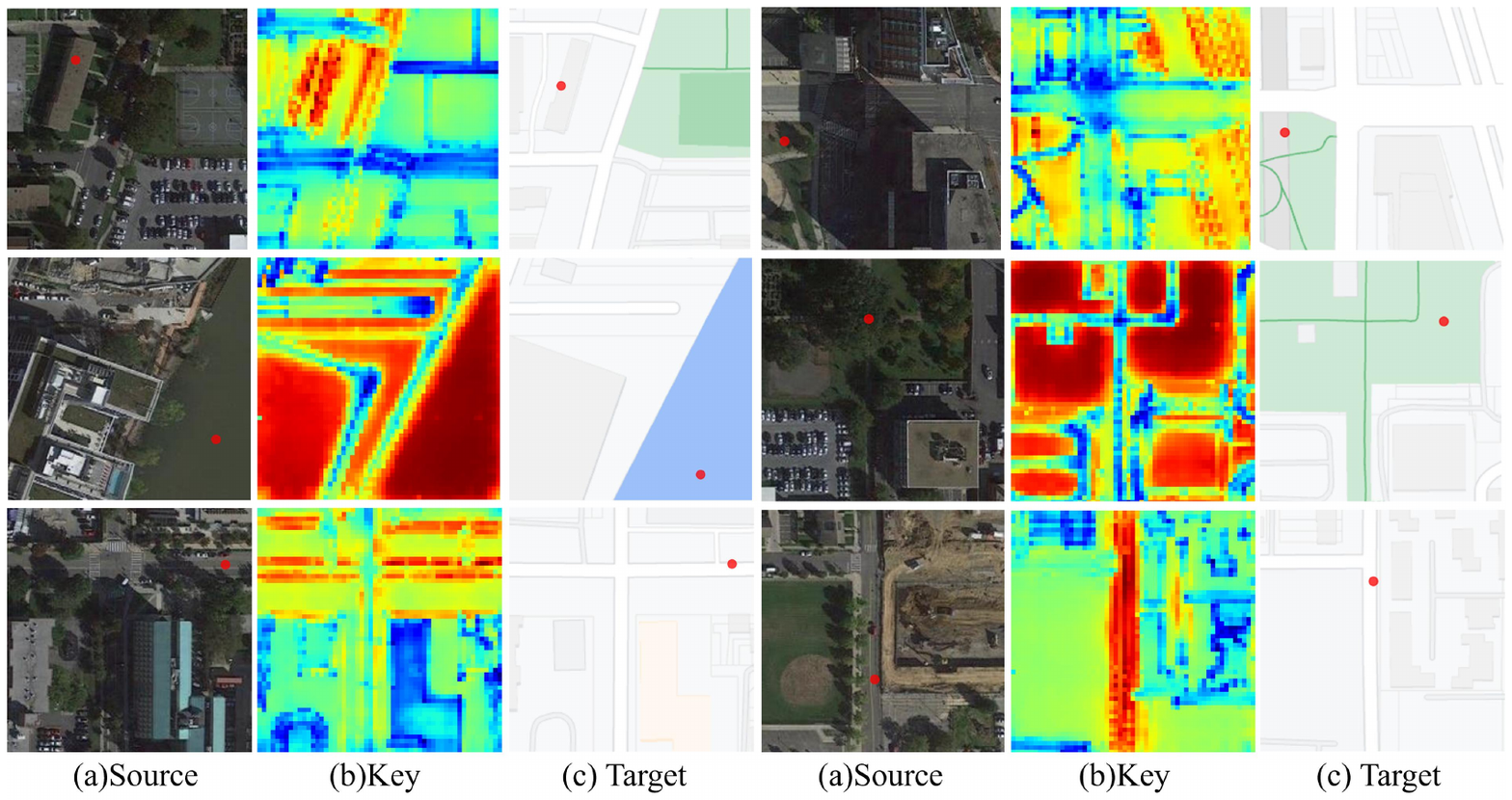}
		\caption{Calculate the similarity between the features related to the red dot in the original image (a) and all the features in the target image (c). Obtain the similarity map (b) by passing $\rm K^L$ in (3) and calculating.}
		\label{fig:10}
	\end{figure}

	Through sections 4.2 and 4.3, the results of our qualitative and quantitative metrics reflect the ability of our map feature perception metrics to have the overall characteristics of the map and the spatial similarity of the map. However, this result may not be intuitive. We thus performed a feature correlation analysis of the original remote sensing images and their corresponding real map images. Specifically, we calculate the similarity of the corresponding feature points (labeled by position vectors) in the original remote sensing image to all features in the target map. As illustrated in Fig. \ref{fig:10}, the semantic features of the remote sensing image demonstrate a high degree of consistency with the map's features for location. Additionally, a degree of correlation is observed at other locations. As can be seen from the correlation map, our indicator can perceive semantics. It can establish a connection between the semantic features in the remote sensing image and the features of the elements in the map; on the other hand, it also successfully identifies the elements in the map
	
	\section{Conclusion}
	This paper proposes a novel evaluation metric for map generation tasks: the map feature perception metric. This metric evaluates the differences between generated and real maps from deep semantic-level features. The experimental results show that after using this index as the loss function, the quality of the generated map is significantly improved. This further verifies that the map feature perception metric can effectively capture and optimize the key features of the map, thereby improving the overall performance of the map generation task.
	\par
	Our study also provides a new perspective on map generation from the semantic and spatial features perspective, emphasizing the importance of thoroughly considering the global structure and semantic relationships unique to maps in the map generation task. This method is significantly innovative in theory and shows its potential in practical applications. For example, by improving the quality of map generation, the navigation accuracy of autonomous driving, the simulation effect of urban planning, and the efficiency and accuracy of map drawing can be further improved. The needs in these fields give our research-wide application value.
	\par
	Future research work can be carried out in the following directions: further optimizing the map feature perception loss to improve its adaptability to more diverse datasets and generative models; exploring its integration with other advanced image generation techniques (such as diffusion models and generative adversarial networks); and attempting to introduce multimodal data (such as text, geographical information, etc.) to expand the applicability of the method. Concurrently, it would be advantageous to investigate the potential of research on map generation methods in interdisciplinary fields, such as disaster emergency management or environmental visualization. In summary, our approach provides novel concepts and instruments for map generation tasks and is expected to encourage further development in this field and have a substantial impact on related practical applications.
	\bibliographystyle{cas-model2-names}
	\bibliography{ref.bib}
	
	
	
*\end{document}